\documentclass[pdflatex,sn-mathphys-num]{sn-jnl}

\let\orcidlogo\relax
\usepackage{orcidlink}
\usepackage{amsmath}
\usepackage{amssymb}
\usepackage{booktabs}
\IfFileExists{siunitx.sty}{\usepackage{siunitx}}{\newcommand{\num}[1]{##1}}
\usepackage{graphicx}
\usepackage{subcaption}
\usepackage{enumitem}
\usepackage{hyperref}
\usepackage{url}

\begin{document}

\title[LoRA Rank Trade-offs in Diffusion Fine-Tuning]{Understanding LoRA Rank Trade-offs in Diffusion Model Fine-Tuning}

\author*[1]{\fnm{Iman} \sur{Khazrak \orcidlink{0000-0001-7087-2283}}}\email{ikhazra@bgsu.edu}
\author[2]{\fnm{Narges} \sur{Nejad \orcidlink{0000-0001-7682-7010}}}\email{narges.nejad@angelo.edu}

\author[1]{\fnm{Mostafa} \sur{M. Rezaee \orcidlink{0000-0002-0849-3650}}}\email{mostam@bgsu.edu}
\author[1]{\fnm{Robert} \sur{C. Green II \orcidlink{0000-0002-3792-2725}}}\email{greenr@bgsu.edu}

\affil[1]{\orgdiv{Department of Computer Science}, \orgname{Bowling Green State University}, \city{Bowling Green}, \state{OH}, \country{USA}}
\affil[2]{\orgdiv{Department of Management and Marketing}, \orgname{Angelo State University}, \orgaddress{\city{San Angelo}, \state{TX}, \country{USA}}}


\abstract{
Selecting LoRA rank for diffusion fine-tuning requires balancing quality and
compute cost. We present a controlled study on CIFAR-10 using a DDPM U-Net
with ranks \(\{2,4,8,16,32\}\), fixed optimization settings, and a reproducible
local-folder \texttt{pytorch-fid} protocol. We report FID, trainable parameters,
runtime, and GPU memory, then validate trends with extended-budget DDPM runs
(20 epochs; ranks 4/8/16) and a Tiny DiT backbone (10 epochs; ranks 4/8/16).
Results show moderate ranks are most efficient: rank 4 achieves the best DDPM
FID (\num{124.1380}), rank 8 is close (\num{124.2136}), and higher ranks provide
limited gains despite larger adaptation cost. These findings support
small-to-moderate ranks as practical defaults under fixed training budgets.
}


{\let\newpage\relax\maketitle}
\noindent\textbf{Submission Type:} Regular Research Paper
  
\newpage

  \section{Introduction}
  \label{sec:intro}

Diffusion models have emerged as a compelling foundation for image generation, achieving
strong performance across a range of visual synthesis tasks \cite{ho2020ddpm,
peebles2023dit}. In many practical deployment scenarios, however, the goal is not to train
a diffusion model from scratch but to adapt a pretrained backbone to a new data
distribution, style, or domain. End-to-end fine-tuning of full diffusion backbones remains
computationally demanding: updating all parameters requires maintaining gradients and
optimizer states for every weight in the network, which quickly becomes prohibitive under
limited compute budgets. This has motivated a growing interest in parameter-efficient
fine-tuning (PEFT) methods, which constrain adaptation to a small, structured subset of
parameters while leaving the pretrained backbone largely intact.

Among available PEFT approaches, Low-Rank Adaptation (LoRA) has become particularly
prevalent in diffusion fine-tuning workflows \cite{hu2022lora}. Its appeal stems from
several practical properties: it requires no changes to the inference architecture, trained
adapters can be merged directly into the base weights, and the adaptation budget is
controlled by a single interpretable hyperparameter---the rank $r$---that determines the
expressivity of the low-rank weight update. These properties make LoRA easy to adopt in
existing pipelines and straightforward to reason about from a systems perspective.

Despite its widespread use, the relationship between LoRA rank and fine-tuning quality
in diffusion models is not well characterized under controlled conditions. In practice,
rank selection is often treated informally: practitioners may default to a fixed rank
borrowed from language model fine-tuning literature, or sweep a small range of values
without systematic reporting. This informal treatment obscures an important empirical
question: under a fixed training budget with uniform optimization settings, how does rank
actually affect adaptation quality, parameter efficiency, memory, and runtime? The answer
is non-obvious. Very low ranks limit the expressivity of the adapter, potentially
underfitting the target distribution. Conversely, unnecessarily large ranks increase
trainable parameter counts and memory pressure without guaranteed quality gains, and may
in fact degrade performance if the optimizer cannot effectively utilize the expanded
capacity within the available budget.

Existing diffusion literature provides limited controlled guidance on this question.
Prior work has largely focused on model quality at scale, novel architectural designs, or
improved sampling procedures \cite{ho2020ddpm, peebles2023dit}, rather than on the
practical engineering problem of rank selection under reproducible and resource-constrained
conditions. Recent work on adaptive rank allocation \cite{zhang2023adalora,
chang2025elalora} has demonstrated that layer-wise rank heterogeneity can improve
efficiency, and rank-stabilized scaling \cite{kalajdzievski2023rslora} has shown that
standard LoRA scaling can impede learning at higher ranks. These contributions are
valuable, but they introduce additional complexity that makes it difficult to isolate the
baseline effect of rank under a simple, uniform training regime. A controlled empirical
study of static rank selection---one that holds all other variables fixed---remains a
useful and underserved reference point for practitioners.

This paper addresses that gap through a controlled empirical study of LoRA rank
trade-offs in diffusion fine-tuning. We conduct a primary rank sweep over
$r \in \{2, 4, 8, 16, 32\}$ using a DDPM U-Net backbone on CIFAR-10, reporting quality
(FID), trainable parameter counts, wall-clock runtime, and GPU memory under fixed
optimization settings. We then assess whether the observed trends are robust to two
sources of variation: additional training budget (extended 20-epoch DDPM runs at ranks
$\{4, 8, 16\}$) and architectural differences (a lightweight Tiny DiT backbone at ranks
$\{4, 8, 16\}$ for 10 epochs). Throughout, we use a local \texttt{pytorch-fid} protocol
with a fixed CIFAR-10 test reference to ensure evaluation consistency across all
experimental conditions. The study is intentionally scoped as a controlled engineering
analysis rather than a claim of state-of-the-art performance: optimization settings are
held fixed across ranks to isolate the effect of rank as a systems variable, and all
findings are interpreted accordingly.
  
  This work makes the following contributions:
  \begin{itemize}[leftmargin=1.5em]
    \item A controlled DDPM LoRA rank sweep on CIFAR-10 across ranks $\{2, 4, 8, 16, 32\}$
    with unified training and evaluation protocols.
    \item Quantitative characterization of efficiency-quality trade-offs using trainable
    parameter counts, runtime, GPU memory, and FID as a relative optimization indicator
    under constrained budgets.
    \item Extended-budget DDPM validation (20 epochs; ranks $\{4, 8, 16\}$) examining
    whether higher ranks close the quality gap with additional training steps.
    \item A Tiny DiT second-backbone validation (10 epochs; ranks $\{4, 8, 16\}$) assessing
    whether rank-ordering trends generalize beyond the DDPM U-Net architecture.
    \item A reproducible PEFT diffusion experimentation pipeline with explicit artifact
    generation, scriptable validation, and structured reporting outputs.
  \end{itemize}

\begin{figure*}[!t]
\centering
\includegraphics[width=\textwidth]{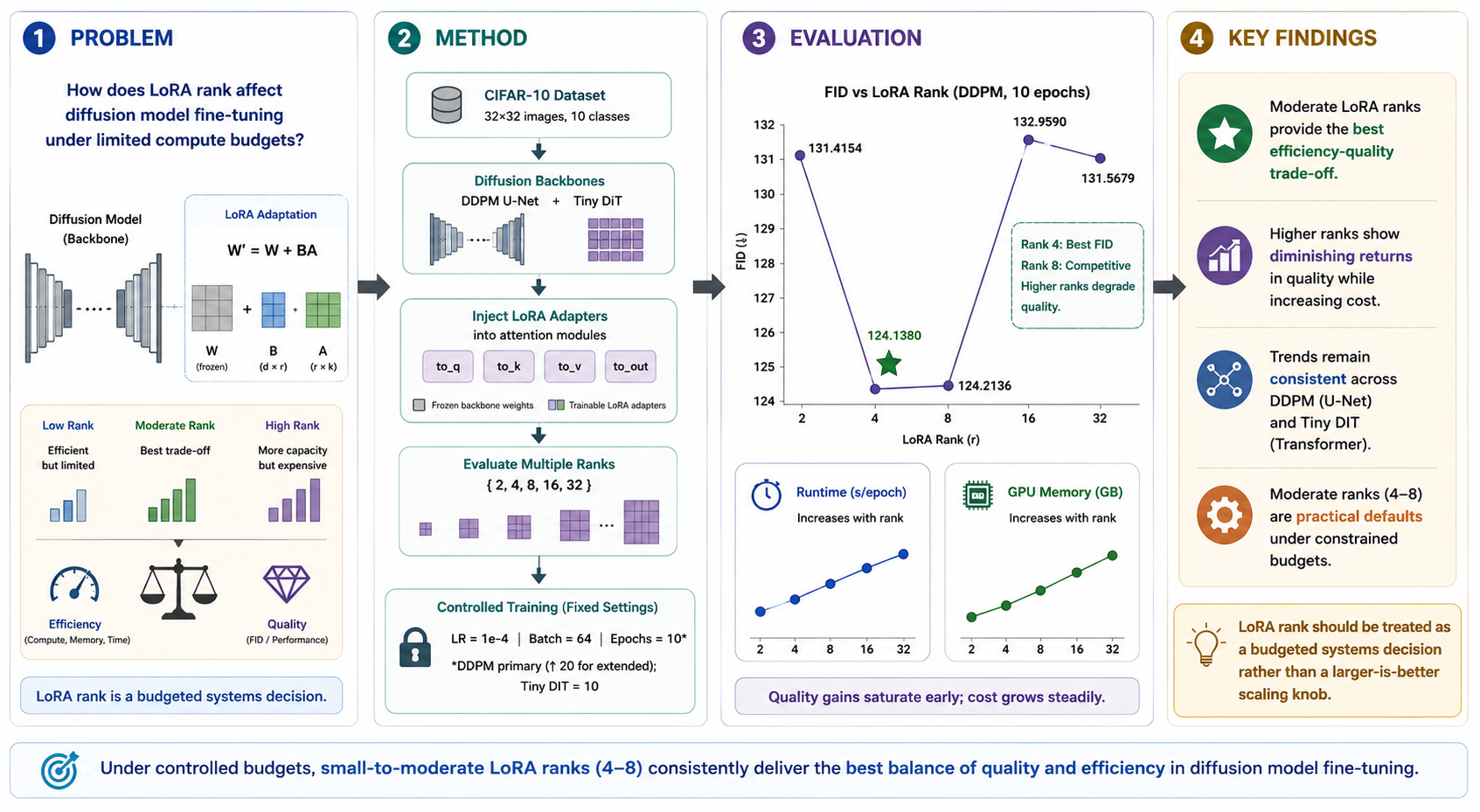}
\caption{
Overview of the controlled evaluation framework for LoRA rank trade-offs in diffusion model fine-tuning. Multiple LoRA ranks are evaluated across DDPM and Tiny DiT backbones under fixed optimization budgets, showing that moderate ranks provide the most favorable efficiency--quality balance.
}
\label{fig:graphical_abstract}
\end{figure*}

  Figure~\ref{fig:graphical_abstract} provides an overview of the experimental workflow and the primary findings of the study. The results show that moderate LoRA ranks consistently provide the strongest efficiency--quality trade-off under fixed optimization budgets.

The remainder of this paper is organized as follows. Section~2 reviews related work on diffusion models and parameter-efficient fine-tuning. Section~3 introduces the problem formulation and technical background. Section~4 presents the methodology and controlled experimental design. Section~5 describes the implementation details and experimental setup. Section~6 reports the results and discussion across all evaluation tracks. Finally, Sections~7 and~8 discuss limitations and conclude the paper.
  
  \section{Related Work}
  \label{sec:related}
  
  Denoising diffusion probabilistic models (DDPMs) establish a likelihood-based generative
  framework in which a neural network learns to iteratively reverse a fixed Gaussian noising
  process applied to training data \cite{ho2020ddpm}. Since their introduction, diffusion
  models have expanded considerably in both scale and architectural diversity, with
  transformer-based backbones such as the Diffusion Transformer (DiT) demonstrating strong
  performance on class-conditional image generation \cite{peebles2023dit}. The present work
  does not propose a new diffusion architecture; rather, it examines practical fine-tuning
  behavior under fixed training budgets, using established backbones as controlled
  experimental platforms.
  
  Parameter-efficient fine-tuning (PEFT) methods reduce adaptation cost by confining
  gradient updates to a small subset of model parameters. Early approaches introduced
  lightweight adapter modules inserted between frozen transformer layers
  \cite{houlsby2019adapters}, while prompt-based tuning methods condition frozen models
  through learned input representations \cite{lester2021prompt}. LoRA offers a complementary
  formulation: rather than inserting new modules, it augments frozen weight matrices with
  trainable low-rank factors that can be folded back into the base weights at inference
  \cite{hu2022lora}. The rank parameter $r$ provides direct, interpretable control over
  adaptation capacity, making LoRA a natural candidate for systematic efficiency-quality
  trade-off analysis under fixed compute budgets.
  
  A growing body of work has examined the limitations of static, uniform rank assignment in
  LoRA. AdaLoRA \cite{zhang2023adalora} reparameterizes weight updates via singular value
  decomposition and dynamically prunes less important singular values based on layer-wise
  importance scores, allowing rank to adapt to the varying demands of different modules.
  Subsequent methods have extended this direction: ElaLoRA \cite{chang2025elalora} enables
  both rank pruning and expansion during training via gradient-derived importance estimates,
  while ARD-LoRA \cite{ardlora2025} targets heterogeneous adaptation requirements across
  individual attention heads. Separately, rsLoRA \cite{kalajdzievski2023rslora} identifies a
  theoretical limitation in the standard LoRA scaling factor, showing that the conventional
  $\alpha/r$ normalization causes gradient magnitude to collapse as rank increases, and
  proposes a corrected $\alpha/\sqrt{r}$ scaling that stabilizes learning at higher ranks
  without additional inference cost. More recently, timestep-dependent rank adaptation has
  been explored for diffusion model personalization: T-LoRA \cite{soboleva2025tlora}
  demonstrates that higher diffusion timesteps are disproportionately prone to overfitting,
  motivating rank schedules that vary with the denoising trajectory.
  
  While these adaptive and rank-aware methods represent important advances, they introduce
  additional complexity---in the form of dynamic schedules, importance scoring, or modified
  optimization regimes---that can obscure the baseline relationship between fixed rank and
  adaptation quality. Our study deliberately occupies a complementary position: rather than
  proposing a new rank-selection mechanism, we provide controlled empirical evidence of how
  static rank choices interact with training budget under uniform optimization settings. This
  evidence is directly relevant for practitioners who require reproducible baselines before
  committing to more complex adaptive strategies.
  
  Open-source frameworks have substantially lowered the barrier to reproducible diffusion experimentation. Hugging Face Diffusers~\cite{vonplaten2022diffusers} provides consistent implementations of models and noise schedulers across a wide range of architectures, enabling fair cross-study comparisons. Reproducible image-quality evaluation commonly relies on the Fr\'echet Inception Distance (FID)~\cite{heusel2017fid}, while CIFAR-10 remains a widely used controlled benchmark that facilitates rapid and resource-efficient comparative studies~\cite{krizhevsky2009cifar}. Recent studies have also explored diffusion-based generative modeling in medical imaging applications, including augmentation for small and imbalanced datasets~\cite{khazrak2025addressing} and synthetic-image-assisted vocal fold pathology classification using DDPM-generated images~\cite{khazrak2025feasibility}. The present study builds directly on these foundations, employing a local \texttt{pytorch-fid} protocol with a fixed CIFAR-10 test reference to ensure evaluation consistency across all experimental tracks.
  
  \section{Problem Statement and Background}
  \label{sec:background}
  
  Selecting the LoRA rank for diffusion fine-tuning is, in essence, a budgeted systems
  decision: the practitioner must allocate a fixed parameter and compute budget across
  fine-tuning runs without per-rank hyperparameter retuning, and must do so in a way that
  is reproducible and transferable to similar settings. Despite the prevalence of LoRA in
  diffusion adaptation workflows, the empirical relationship between rank and adaptation
  quality under such constraints is not well characterized. This section introduces the
  necessary technical background and clarifies the design implications that motivate the
  controlled evaluation in this paper.
  
  Denoising diffusion probabilistic models learn to generate data by reversing a fixed
  forward process that progressively corrupts clean samples with Gaussian noise. Formally,
  the model trains a neural network $\epsilon_\theta(x_t, t)$ to predict the noise injected
  at diffusion step $t$, where $x_t$ is obtained by sampling from the forward noising process
  applied to a clean input $x_0$ \cite{ho2020ddpm}. At inference time, new samples are
  generated by iteratively applying the learned denoising network from a pure Gaussian prior,
  guided by a fixed noise schedule. The quality of this reverse process depends critically on
  the capacity of $\epsilon_\theta$ to model the score function of the data distribution
  across all noise levels, a requirement that motivates the use of expressive U-Net and
  transformer-based backbone architectures \cite{ho2020ddpm, peebles2023dit}.
  
  LoRA adapts a pretrained weight matrix $W \in \mathbb{R}^{d_{\text{out}} \times d_{\text{in}}}$
  by introducing a low-rank residual update of the form
  \begin{equation}
    W' = W + BA,
    \label{eq:lora}
  \end{equation}
  where $A \in \mathbb{R}^{r \times d_{\text{in}}}$ and
  $B \in \mathbb{R}^{d_{\text{out}} \times r}$ are the trainable adapter matrices, and
  $r \ll \min(d_{\text{out}}, d_{\text{in}})$ is the rank \cite{hu2022lora}. During
  fine-tuning, the base parameters $W$ are frozen and only $A$ and $B$ are optimized,
  substantially reducing the number of trainable parameters relative to full fine-tuning.
  The total adapter parameter count scales linearly with $r$: increasing rank expands the
  expressivity of the low-rank update but proportionally increases trainable parameter
  counts, memory footprint, and, to a lesser extent, runtime. Crucially, the adapted weight
  $W'$ takes the same form as the original matrix, so trained adapters can be merged into
  $W$ at inference with no additional computational overhead.
  
  Because rank directly governs trainable capacity, comparing LoRA configurations across
  different ranks is a non-trivial experimental design problem. If optimizer settings are
  adjusted separately for each rank, the observed quality differences conflate the effect of
  rank with the effect of tuning. Our study therefore holds all optimization hyperparameters
  fixed across ranks, interpreting observed FID differences as evidence of rank efficiency
  under a controlled and uniform training budget, rather than as indicators of per-rank
  hyperparameter optima. This design choice is central to the empirical framing of all
  results reported in Section~\ref{sec:results}.
  
  \section{Methodology}
  \label{sec:methodology}
  
  \subsection{Study Design}
  
  The study is organized into three controlled experiment tracks. The primary track conducts
  a full DDPM rank sweep over $r \in \{2, 4, 8, 16, 32\}$ for 10 epochs, establishing the
  baseline efficiency-quality profile across the evaluated rank range. The second track
  extends the training budget of the DDPM configuration to 20 epochs for a subset of ranks
  $r \in \{4, 8, 16\}$, testing whether higher ranks recover quality advantages with
  additional optimization steps. The third track applies the same 10-epoch protocol to a
  lightweight Tiny DiT backbone at ranks $r \in \{4, 8, 16\}$, assessing whether the
  rank-ordering trends observed in the primary DDPM sweep generalize to a structurally
  different architecture. All three tracks use CIFAR-10 at $32 \times 32$ resolution and
  share identical core optimization settings: batch size 64 and learning rate $10^{-4}$.
  
  \subsection{LoRA Injection and Parameter Freezing}
  
  LoRA adapters are injected into the attention projection modules of each target backbone,
  specifically the query, key, value, and output projections (\texttt{to\_q}, \texttt{to\_k},
  \texttt{to\_v}, \texttt{to\_out.0}). All base model parameters are frozen throughout
  training, and only the LoRA adapter matrices are updated via gradient descent. For each
  experimental run, we record both total and trainable parameter counts directly from run
  artifacts, allowing precise verification that parameter scaling across ranks matches the
  theoretical expectation implied by Equation~\ref{eq:lora}.
  
  \subsection{Evaluation Protocol}
  
  Generation quality is assessed using the Fr\'{e}chet Inception Distance
  (FID)~\cite{heusel2017fid}, computed via \texttt{pytorch-fid} against a fixed local
  CIFAR-10 test reference set stored at \texttt{outputs/fid\_reference/cifar10\_test}. To
  maintain evaluation consistency across all tracks and ranks, FID is computed from 2000
  generated images per run. Using a local reference folder rather than a precomputed dataset
  statistic eliminates a common source of cross-study evaluation drift. In addition to FID,
  we record final training loss, wall-clock runtime, and peak GPU memory consumption from
  training artifacts, providing a joint view of quality and systems efficiency for each rank
  configuration.
  
  \subsection{Controlled-Optimization Positioning}
  
  A central design decision in this study is the deliberate choice to hold all optimization
  settings fixed across ranks. This means that differences in FID across ranks reflect the
  effect of rank under a uniform training budget, rather than the outcome of rank-specific
  hyperparameter search. As a consequence, results should be interpreted as controlled
  within-budget comparisons: they characterize which rank configurations make the most
  efficient use of a fixed compute allocation, not which ranks would perform best given
  tailored optimization. This framing is carried consistently through all experimental tracks
  and is fundamental to the interpretation of the findings in Section~\ref{sec:results}.
  
  \section{Experimental Setup}
  \label{sec:experiments}
  
  \subsection{Environment and Stack}
  
  Experiments were executed on OSC via SLURM with GPU-backed jobs. The implementation stack
  uses PyTorch, Hugging Face Diffusers \cite{vonplaten2022diffusers}, PEFT/LoRA
  \cite{hu2022lora}, and Hugging Face datasets tooling. All runs are generated through
  scripted pipelines that produce structured artifacts at each stage, including per-rank
  training logs, generated image sets, and FID evaluation outputs, supporting end-to-end
  reproducibility.
  
  \subsection{Data and Backbones}
  
  All experiments use CIFAR-10 \cite{krizhevsky2009cifar} at $32 \times 32$ resolution as
  the training and evaluation dataset. The primary backbone is a DDPM U-Net
  \cite{ho2020ddpm}, selected for its widespread use in controlled diffusion fine-tuning
  studies. The validation backbone is a lightweight Tiny DiT-style
  \texttt{Transformer2DModel} inspired by the DiT architecture \cite{peebles2023dit},
  included as a cross-backbone check rather than as a large-scale benchmark. The two
  backbones differ in architectural family---convolutional versus transformer-based---allowing
  a limited assessment of whether rank-ordering trends are backbone-specific or more broadly
  consistent.
  
  \subsection{Training Configuration and Reported Metrics}
  
  Across all tracks, the optimizer policy is held fixed: batch size 64, learning rate
  $1 \times 10^{-4}$, with only rank-specific LoRA adapter parameters trainable. The main
  DDPM sweep and Tiny DiT validation run for 10 epochs; the extended-budget DDPM validation
  runs for 20 epochs. For each run, we report FID (computed via \texttt{pytorch-fid} against
  the local CIFAR-10 test reference using 2000 generated images), trainable parameter count
  and percentage of total parameters, wall-clock runtime in seconds, peak GPU memory usage
  in megabytes, and final training loss. This joint reporting of quality and systems metrics
  is intended to provide a complete view of the efficiency-quality trade-off at each rank.
  
  \section{Results and Discussion}
  \label{sec:results}
  
  \subsection{Main DDPM Rank Sweep}
  \label{sec:main_results}
  
  Table~\ref{tab:main_ddpm} summarizes the main controlled DDPM sweep across ranks
  $\{2, 4, 8, 16, 32\}$. Rank 4 achieves the most favorable FID (\num{124.1380}), while
  rank 8 is nearly identical (\num{124.2136}). Higher ranks (16 and 32) increase trainable
  parameter budgets substantially but do not improve FID under the same training budget,
  indicating diminishing returns in adaptation quality relative to parameter cost.
  
  \begin{table}[t]
    \centering
    \caption{Main DDPM sweep (10 epochs, CIFAR-10, fixed optimization settings).}
    \label{tab:main_ddpm}
    \small
    \begin{tabular}{rrrrrrr}
      \toprule
      Rank & Trainable & Trainable(\%) & Runtime(s) & Peak MB & Final Loss & FID \\
      \midrule
      2  & 49152  & 0.0432 & 521 & 2048.21 & 0.0532 & 131.4154 \\
      4  & 98304  & 0.0864 & 522 & 2048.89 & 0.0594 & 124.1380 \\
      8  & 196608 & 0.1727 & 524 & 2050.24 & 0.0587 & 124.2136 \\
      16 & 393216 & 0.3447 & 521 & 2052.95 & 0.0414 & 132.2549 \\
      32 & 786432 & 0.6871 & 528 & 2074.49 & 0.0634 & 131.6479 \\
      \bottomrule
    \end{tabular}
  \end{table}
  
  \subsection{Efficiency--Quality Trade-offs}
  \label{sec:efficiency}
  
\begin{figure}[t]
    \centering
    
    \begin{minipage}{0.49\linewidth}
        \centering
        \includegraphics[width=\linewidth]{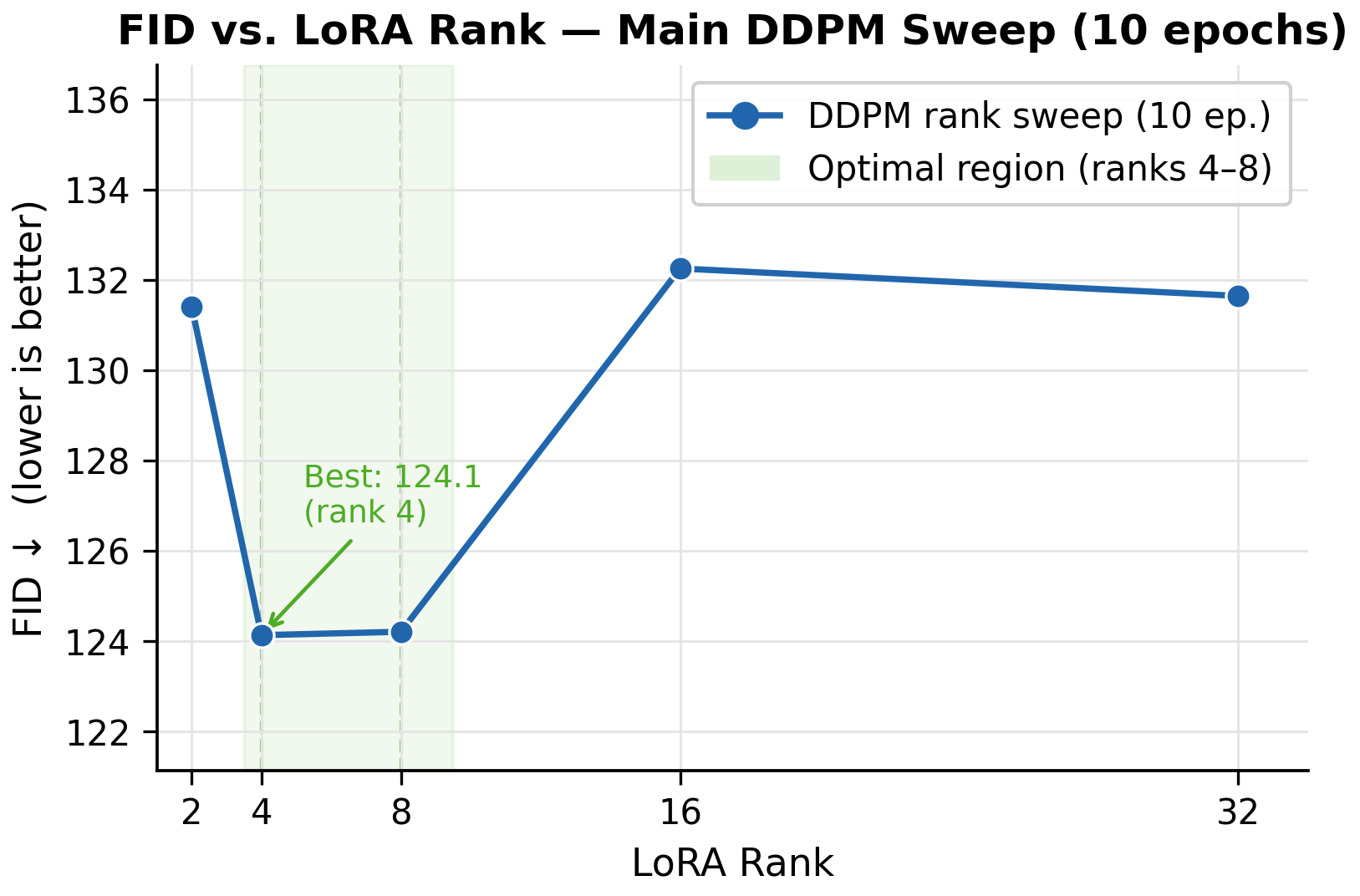}
        \caption{Main DDPM sweep: FID versus LoRA rank.}
        \label{fig:fid_vs_rank}
    \end{minipage}
    \hfill
    \begin{minipage}{0.49\linewidth}
        \centering
        \includegraphics[width=\linewidth]{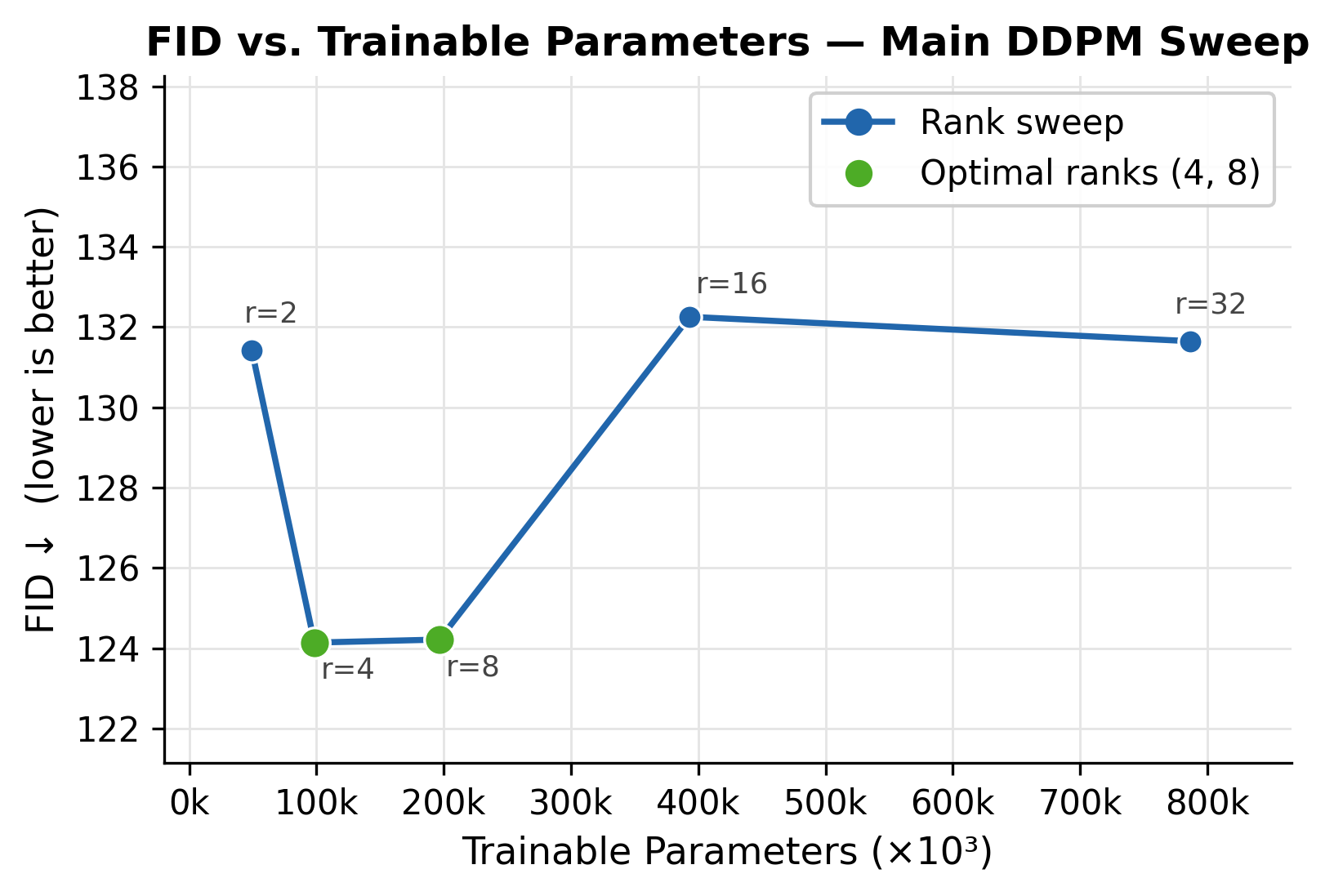}
        \caption{Main DDPM sweep: FID versus trainable parameters.}
        \label{fig:fid_vs_trainable}
    \end{minipage}
    
\end{figure}

  Figure~\ref{fig:fid_vs_rank} shows FID as a function of rank in the main sweep. The
  relationship is non-monotonic: quality improves from rank 2 to rank 4, plateaus at rank 8,
  and then degrades at ranks 16 and 32. This pattern suggests that, under the fixed
  optimization budget used here, moderate ranks provide the best balance between adaptation
  expressivity and optimization tractability. Figure~\ref{fig:fid_vs_trainable} reinforces
  this point, illustrating that increasing trainable parameters does not yield monotonic FID
  improvement under fixed budgets.
  
  

  Runtime and memory overhead scale modestly but consistently with rank (Table~\ref{tab:main_ddpm}),
  while quality gains saturate quickly after rank 4. This asymmetry between resource cost
  and quality return supports an engineering recommendation to prioritize moderate ranks when
  training budgets are fixed and tuning time is limited.
  
  \subsection{Extended-Budget Validation}
  \label{sec:extended}
  
  To test whether higher ranks recover quality advantages with more optimization steps, we
  run a 20-epoch DDPM validation on ranks $\{4, 8, 16\}$. Results are shown in
  Table~\ref{tab:extended_ddpm}. Compared with rank 16 at 10 epochs (FID \num{132.2549}),
  rank 16 at 20 epochs improves to \num{131.3555}, an absolute gain of \num{0.8994} and a
  relative improvement of approximately \num{0.6801}\%. The gain is measurable but modest,
  and does not reverse the broader moderate-rank preference observed under controlled budgets.
  
  \begin{table}[t]
    \centering
    \caption{Extended-budget DDPM validation (20 epochs, ranks 4/8/16).}
    \label{tab:extended_ddpm}
    \small
    \begin{tabular}{rrrrrrr}
      \toprule
      Rank & Trainable & Trainable(\%) & Runtime(s) & Peak MB & Final Loss & FID \\
      \midrule
      4  & 98304  & 0.0864 & 1045 & 2065.01 & 0.0407 & 130.8580 \\
      8  & 196608 & 0.1727 & 1044 & 2066.37 & 0.0343 & 132.2700 \\
      16 & 393216 & 0.3447 & 1040 & 2069.07 & 0.0311 & 131.3555 \\
      \bottomrule
    \end{tabular}
  \end{table}
  
\begin{figure}[t]
    \centering
    \includegraphics[width=0.75\linewidth]{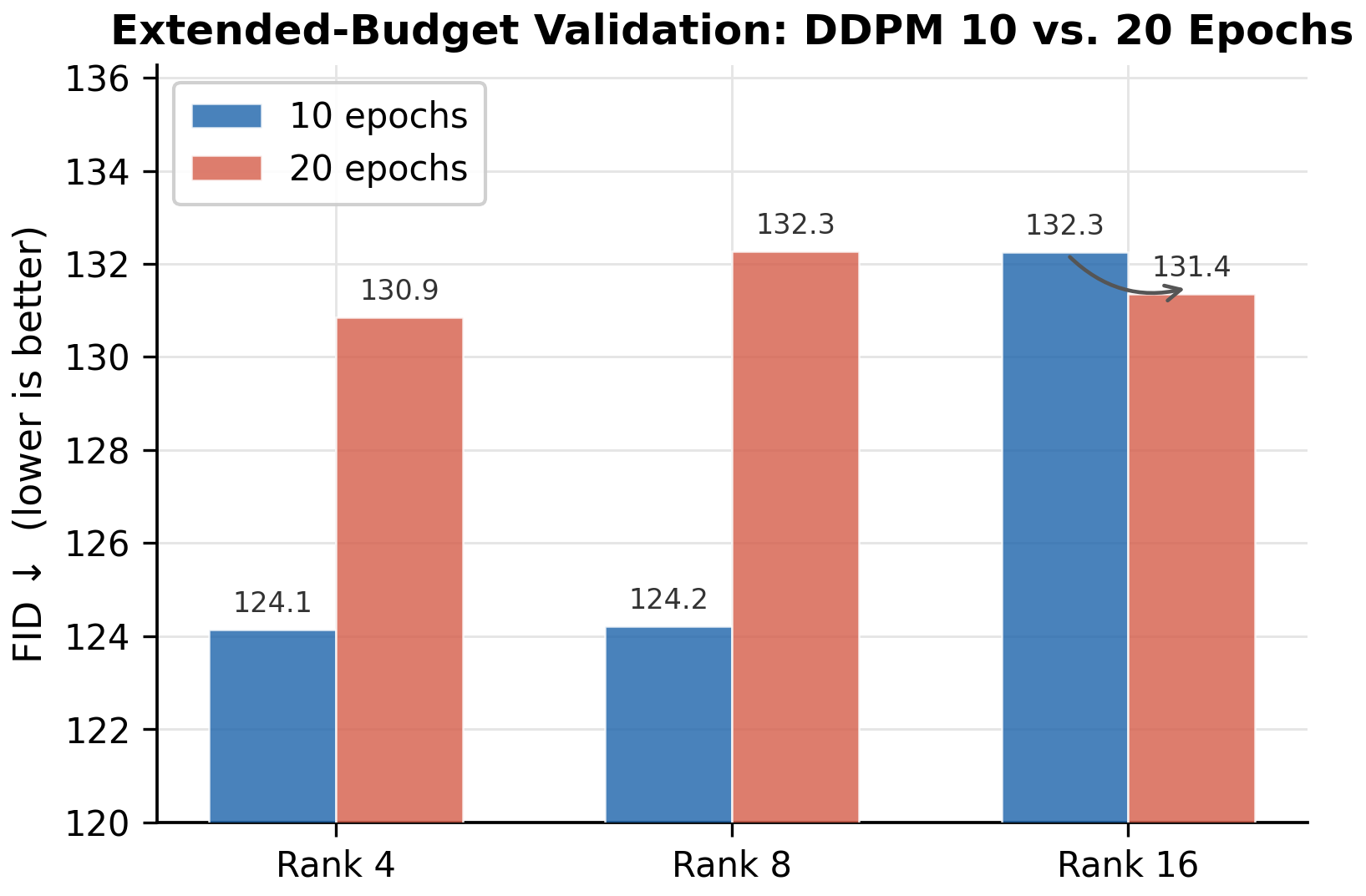}
    \caption{DDPM comparison at 10 vs.\ 20 epochs for ranks 4, 8, and 16.}
    \label{fig:ddpm_10_20}
  \end{figure}

  Figure~\ref{fig:ddpm_10_20} contrasts DDPM performance at 10 and 20 epochs for ranks
  4, 8, and 16, providing a visual comparison of the modest budget-induced gains. Notably,
  rank 4 continues to achieve the lowest FID at 20 epochs (\num{130.8580}), further
  supporting the observation that moderate ranks make more efficient use of available
  optimization steps.
  
  \subsection{Tiny DiT Validation}
  \label{sec:tiny_dit}
  
  To assess whether the rank-ordering trends observed with the DDPM U-Net generalize to a
  structurally different backbone, we evaluate a lightweight Tiny DiT-style
  \texttt{Transformer2DModel} under the same 10-epoch protocol. Results are reported in
  Table~\ref{tab:tiny_dit}. Rank 4 again achieves the best FID (\num{383.2438}), rank 8
  remains competitive (\num{387.8070}), and rank 16 is clearly worse (\num{402.0335}) under
  the same budget. The rank-ordering is consistent with the DDPM findings, despite the
  architectural differences between the two backbones.
  
  \begin{table}[t]
    \centering
    \caption{Tiny DiT validation (10 epochs, ranks 4/8/16).}
    \label{tab:tiny_dit}
    \small
    \begin{tabular}{rrrrrrr}
      \toprule
      Rank & Trainable & Trainable(\%) & Runtime(s) & Peak MB & Final Loss & FID \\
      \midrule
      4  & 16384 & 1.5269 & 549 & 1753.54 & 0.1435 & 383.2438 \\
      8  & 32768 & 3.0078 & 545 & 1761.75 & 0.1425 & 387.8070 \\
      16 & 65536 & 5.8400 & 545 & 1778.18 & 0.1347 & 402.0335 \\
      \bottomrule
    \end{tabular}
  \end{table}
  
\begin{figure}[t]
    \centering
    \includegraphics[width=0.9\linewidth]{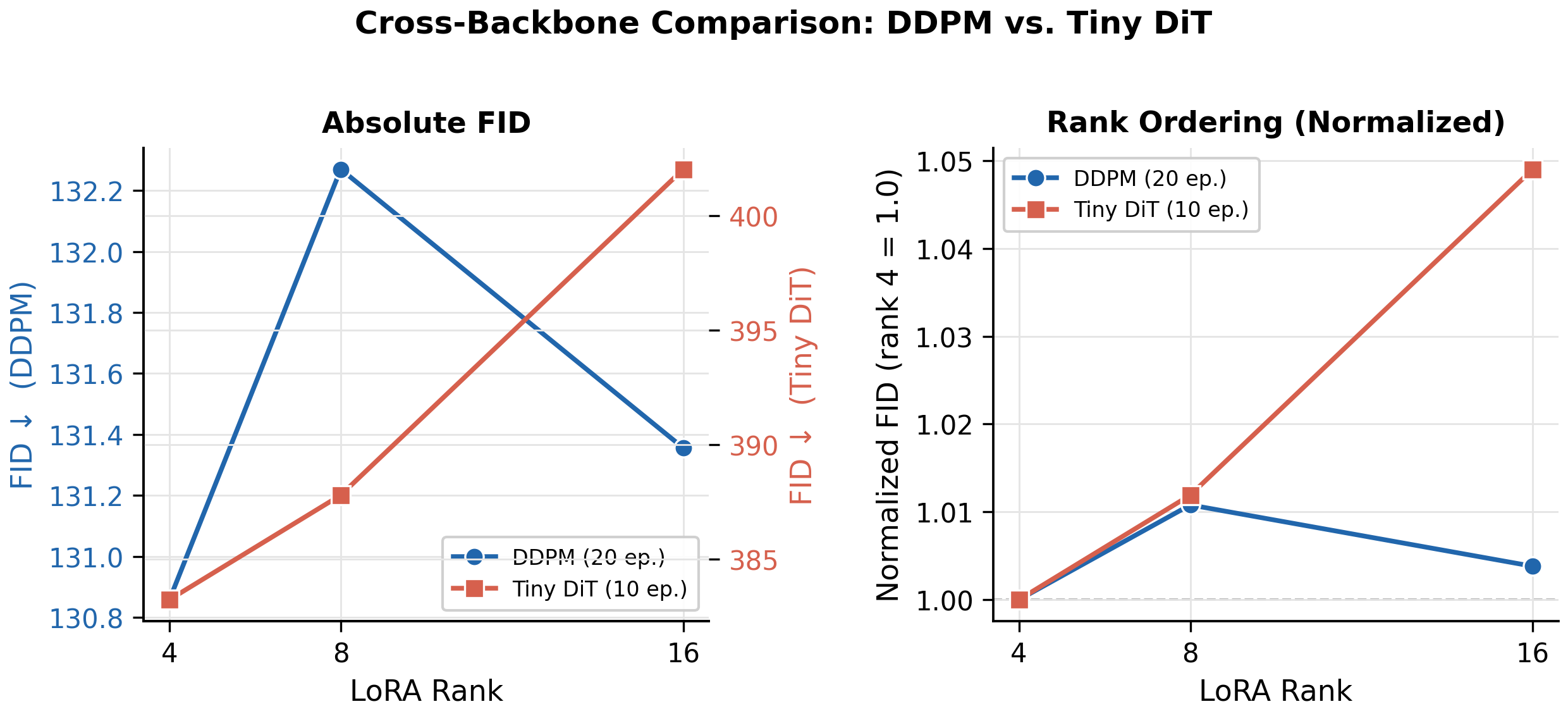}
    \caption{Backbone comparison: DDPM (20 epochs) vs.\ Tiny DiT (10 epochs) for ranks
    4, 8, and 16.}
    \label{fig:ddpm_vs_tiny}
  \end{figure}

  Figure~\ref{fig:ddpm_vs_tiny} compares DDPM (20 epochs) and Tiny DiT (10 epochs) FID
  profiles for ranks 4, 8, and 16. Absolute FID levels are not directly comparable across
  backbones, as the two architectures operate at different quality baselines under the
  evaluated budgets. The key result is the consistency of rank ordering rather than the
  absolute values, which reinforces the generality of the moderate-rank preference within
  the controlled settings of this study.
  
  \subsection{Interpretation and Practical Implications}
  \label{sec:discussion}
  
  Taken together, the results across all three tracks indicate that LoRA rank should be
  treated as a budgeted systems decision rather than a ``larger is always better'' scaling
  knob. In this controlled setting, moderate ranks---especially rank 4, with rank 8 close
  behind---consistently provide stronger quality-efficiency outcomes than larger ranks, and
  this pattern holds across both a 10-epoch and a 20-epoch DDPM setting and across two
  structurally distinct backbones.
  
  An important caveat governs the interpretation of these FID values. All models in this
  study were trained under deliberately constrained budgets on a low-resolution benchmark,
  and the generated samples remained visually undertrained under these conditions. Reported
  FID values should therefore be interpreted primarily as relative optimization indicators
  rather than as measures of final generative quality. The relevant evidence lies in the
  consistency of rank ordering across experimental tracks, not in the absolute FID levels
  achieved.
  
  Under constrained and fixed optimization budgets, increasing rank can expand trainable
  capacity faster than the optimization process can utilize it, yielding weak or even
  negative marginal quality gains. This does not imply that high ranks are universally
  ineffective; rather, it suggests that larger ranks may require different optimization
  budgets and tuning regimes to realize their potential. Within the budget regime studied
  here, however, moderate ranks make more efficient use of available compute.
  
  For practitioners tuning diffusion models under finite compute, these findings suggest
  starting with moderate LoRA ranks before allocating resources to large-rank sweeps, and
  tracking both quality and systems metrics---runtime and memory alongside FID---rather than
  optimizing for FID alone. The study also highlights the value of explicit artifact
  pipelines: rank-specific configurations, local reference data for FID, and scriptable
  validation steps lower the risk of hidden evaluation drift and simplify
  reviewer-facing reproducibility.
  
  \section{Limitations}
  \label{sec:limitations}
  
  This study is intentionally scoped as a controlled empirical engineering analysis, and
  several important limitations should be considered when transferring its conclusions to
  broader settings. All experiments use CIFAR-10 exclusively, at $32 \times 32$ resolution,
  which limits the generalizability of the findings to higher-resolution datasets and more
  diverse data distributions. The evaluated backbones---a DDPM U-Net and a lightweight Tiny
  DiT---are considerably smaller than the large-scale foundation models used in production
  diffusion systems, and rank-ordering trends may differ at scale. Training budgets and
  evaluation sample counts are bounded by reproducibility and resource constraints, and we
  do not perform rank-specific optimizer retuning or exhaustive hyperparameter search. As
  noted in Section~\ref{sec:discussion}, the constrained training budgets result in visually
  undertrained samples, and FID comparisons should be read as relative optimization evidence
  rather than definitive quality rankings. Finally, the results should not be interpreted as
  general scaling laws; they are specific to the optimization regime, architectures, and
  dataset examined here.
  
  \section{Conclusion}
  \label{sec:conclusion}
  
  We presented a reproducible empirical study of LoRA rank trade-offs in diffusion
  fine-tuning under controlled training budgets. In the main DDPM sweep, moderate ranks
  (4 and 8) provided the most favorable quality-efficiency outcomes, while larger ranks
  showed diminishing returns relative to their increased trainable parameter cost.
  Extended-budget DDPM validation improved rank 16 only modestly, and Tiny DiT validation
  exhibited a consistent rank-ordering trend across a structurally different backbone.
  
  These results support a practical recommendation for budget-constrained diffusion
  adaptation: small-to-moderate LoRA ranks are strong default choices when optimization
  settings are fixed and reproducibility is prioritized. Future work should evaluate adaptive
  rank allocation strategies, rank-specific optimizer tuning, larger-scale backbones, and
  broader datasets to determine how these trends evolve under less constrained experimental
  conditions.

\end{document}